\def\year{2018}\relax
\documentclass[letterpaper]{article} 
\usepackage{aaai18}  
\usepackage{times}  
\usepackage{helvet}  
\usepackage{courier}  
\usepackage{url}  
\usepackage{graphicx}  
\usepackage{amsmath}
\usepackage{amssymb,floatrow,multirow,color,algorithm, algorithmic}
\usepackage{graphicx, epsfig, epstopdf, subfigure}

\begin{document}
%
\title{Discriminant Projection Representation-based Classification for Vision Recognition}
\author{Qingxiang Feng and Yicong Zhou*\\
Computer and Information Science,
University of Macau\\
fengqx1988@gmail.com,~yicongzhou@umac.mo,~*Corresponding author.\\
}
\maketitle

\begin{abstract}

Representation-based classification methods such as sparse representation-based
classification (SRC) and linear regression classification (LRC) have attracted
a lot of attentions. In order to obtain the better representation, a novel method called projection
representation-based classification (PRC) is proposed for image recognition in this paper. PRC is based on a new mathematical model.
This model denotes that the 'ideal projection' of a sample point $x$ on the hyper-space $H$ may be gained by iteratively computing
the projection of $x$ on a line of hyper-space $H$ with the proper strategy.
Therefore, PRC is able to iteratively approximate the 'ideal representation' of each subject for classification.
Moreover, the discriminant PRC (DPRC) is further proposed, which obtains the discriminant information by maximizing the ratio of the between-class reconstruction error
over the within-class reconstruction error.
Experimental results on five typical databases show that the proposed PRC and DPRC are effective
and outperform other state-of-the-art methods on several vision recognition tasks.

\end{abstract}


\section{Introduction}

Recently, representation-based classifiers have attracted increasing attentions
of researchers, which can be roughly divided into two kinds: all-classes-based and single-class-based. In the first kind, the well-known method is sparse representation-based classification (SRC) \cite{SRC2}. It was developed to use the all-class-model to obtain the $L_1$-based sparse representation
for classification . To improve the computation efficiency, the collaborative representation-based classification (CRC) \cite{CRC} was proposed to address the $L_2$
minimum problem. Later, several improved methods of SRC were proposed for image recognition \cite{RCR,KCSR,KRDU}, such as
manifold constraint transfer (MCT) \cite{MCT} applies a
strategy to produce new data for classification. Different from the all-class-model in SRC, some classifiers use single
class to obtain its representation. For example, linear regression-based
classification (LRC) \cite{LRC} was proposed for face identification, which was based on that samples from a specific object
class are known to lie on a linear subspace \cite{LRLS,PLRC,ILRC}. LRC solves the least square
errors and obtain the linear projection point as the representation for an independent
class-specific models.

The common objective of existing representation-based methods is to find the best representation for classification. However, they have only obtained a roughly approximated representation. For example,
In ref.\cite{SRC2}, we know that the ideal representation of SRC is to solve the
$L_0$ minimum problem. In LRC, the regression projection is obtained by a matrix's pseudo-inverse.
Therefore, we know that they only obtain the approximated representation.
In order to find a better representation of an image, this paper proposes a
projection representation-based classification (PRC) for image recognition.
To approximate the 'ideal representation' of a sample point, PRC utilizes a
new mathematical model to iteratively compute the projection point of the
test sample towards a line linking a paired of specific points. This
mathematical model has been proved by a theorem. According to
the theorem, we know that the generated projection will be almost equal to
the 'ideal representation' after sufficient iterations.  Moreover, the
discriminant PRC (DPRC) is further proposed, which obtains
the discriminant information by maximizing the ratio of the
between-class reconstruction error over the within-class re-construction error.
The main contributions of this paper are as follows:

\vspace{0.02in}
$\odot$ Firstly, we propose a new mathematical model to obtain the projection of
a point on a hyper-space, and we prove this model mathematically.
\vspace{0.02in}

$\odot$ Secondly, with the mathematical model, we propose the projection
representation-based classification (PRC) for image recognition tasks. The generated projection by PRC will be almost equal to
the 'ideal representation' after sufficient iterations.
\vspace{0.02in}

$\odot$  In order to obtain an effective discriminant subspace for PRC, we
propose the discriminant PRC (DPRC). DPRC utilizes the labeled training samples set to constitute a more reliable subspace such that the effective discriminant information can be used for classification.
\vspace{0.02in}

$\odot$  Experiments have been carried out on several challenging databases. The results show that the proposed PRC and DPRC outperform several state-of-the-art methods.
\\
\\
\textbf{Notation Summary}\\
Let $X=\{x_i^c\in R^{q\times 1},i=1,2,\cdots,N_c,c=1,2,\cdots,M\}$ denote
the prototype data set, where  $x_i^c$ is the $i^{\mbox{th}}$ sample of the $c^{\mbox {th}}$ class, $M$ is
the number of classes, $N_c$  is the number of samples of
the $c^{\mbox{th}}$ class and $q$ is the sample's dimension. The number of all the samples is $L=\sum\limits_{c = 1}^M N_c$. The prototype data set can be also described as
$X=\{x_i\in R^{q\times 1},i=1,2,\cdots,L\}$.

%
%

%


\section{Proposed Math Model }



%

Before introducing the math model, we describe the 'ideal projection' in Definition 1.
\\
\\
\textbf{Definition 1:} \emph{Suppose that there exists a test sample $x$ and a specific class $c$. If a point in the
hyper-space of the $\mbox{c}^{\mbox{th}}$ class is the nearest to test sample $x$, it is
treated as the 'ideal representation' or 'ideal projection' of the test sample $x$ on the hyper-space.}

\subsection{Math Model}

Given a point $x$ and a hyper-space $H$, the 'ideal projection' of $x$ on the
hyper-space $H$ may be gained by iteratively computing the projection point of $x$
on a line of hyper-space $H$ with the proper strategy. It can be describe as
\begin{eqnarray}
p\{x,H\}\approx\underset{k\rightarrow+\infty}{repreat}~~p\{x,\overline{x_i^{c,k}x_{i^*}^{c,k}}\}
\end{eqnarray}
where $p\{*,\bigotimes\}$ denotes the projection of $*$ on $\bigotimes$, $x_i^{c,k}$, $x_{i^*}^{c,k}$
are two points of the hyper-space $H$. $x_{i^*}^{c,k}$ is the nearest point among all known-distance points of $H$.

%

\subsection{Correctness of the Math Model}

Now, we know that the proposed model is quite useful for finding the better representation
of each class for classification. Therefore, the correctness of the proposed math model
will be an important problem. The Theorem 1 is provided to prove the proposed math model.
According to the Theorem 1, a projection point with the minimum distance can be obtained by iteratively computing the projection point of $x$
on hyper-space's a line. Considering Definition 1, we know that the obtained projection point can be treated as
the projection of $x$ on the hyper-space. Therefore, the proposed math model is correct.
\\
\\
\textbf{Theorem 1:}  \emph{Given a test sample $x$, and a specific class $c$ with $N_c$ training
samples.
Suppose the $\mbox{c}^{th}$ class in the first round of projection is
$X_c^0=[x_1^{c,0}~x_1^{c,0}~\cdots~x_{N_c}^{c,0}]$. Select the nearest training sample
$x_n^{c,0}$  and another training sample $x_i^{c,0}$, $i\neq n$  to form a line
$\overline{x_n^{c,0}x_i^{c,0}}$; Compute the projection point $x_p^{c,0}$ of the
test sample $x$ to the line $\overline{x_n^{c,0}x_i^{c,0}}$; and use $x_p^{c,0}$ to
replace $x_i^{c,0}$, $i\neq n$  as the new training set of the $c^{th}$ class. If
this projection operation was performed in sufficient times, the distance between the
test sample and new projection point closely approximates to a fixed constant,
which is the smallest distance between the test sample $x$ and the space of the $c^{th}$ class.}
\\
\\
\textbf{Proof:}  Because $x_p^{c,0}$ is the projection of the test sample $x$ to the line
$\overline{x_n^{c,0}x_i^{c,0}}$, then
\begin{eqnarray}
\|x-x_p^{c,0}\|\leq\|x-x_n^{c,0}\|.
\end{eqnarray}
After the first projection procedure, $x_p^{c,0}$ replaces $x_n^{c,0}$ as the
new nearest sample and will be used to form the new line. Following this manner,
in the $k^{th}$ projection procedure, we have
\begin{eqnarray}
\|x-x_p^{c,k}\|\leq \|x-x_p^{c,k-1}\|
\end{eqnarray}
Because the distance between the test sample $x$ and the projection point is equal
or greater than 0, the projection points satisfy the following conditions.
\begin{eqnarray}
0\leq\|x-x_p^{c,k}\|\leq\|x-x_p^{c,k-1}\|,~k=1,2,\cdots,+\infty
\end{eqnarray}
That is,
\begin{eqnarray}
\lim\limits_{k\rightarrow+\infty}\|x-x_p^{c,k}\|=d
\end{eqnarray}
where $d$ is a constant that is the smallest distance between the test sample $x$
and the subspace of the $c^{th}$ class.

\subsection{Proposed Math Model vs Linear Regression}

For a specific class subspace,the real projection point cannot be computed using the existing math knowledge because the class subspace is a hyper-space.
 Ref. \cite{LRC} proposed LRC to solve the least square errors and obtain the linear projection point. LRC has the good performance. However, LRC obtains the linear projection point by a pseudo-inverse operation such that this point is only a roughly approximated projection point (exist the closer
point than the linear projection point), not the ideal projection point according to Definition 1. Therefore,
we intend to obtain a better projection point that is the nearest one to the 'ideal projection' point
by using the proposed math model.

\section{Proposed PRC}
\begin{figure}[t]
\begin{center}
\includegraphics[width=0.75\linewidth]{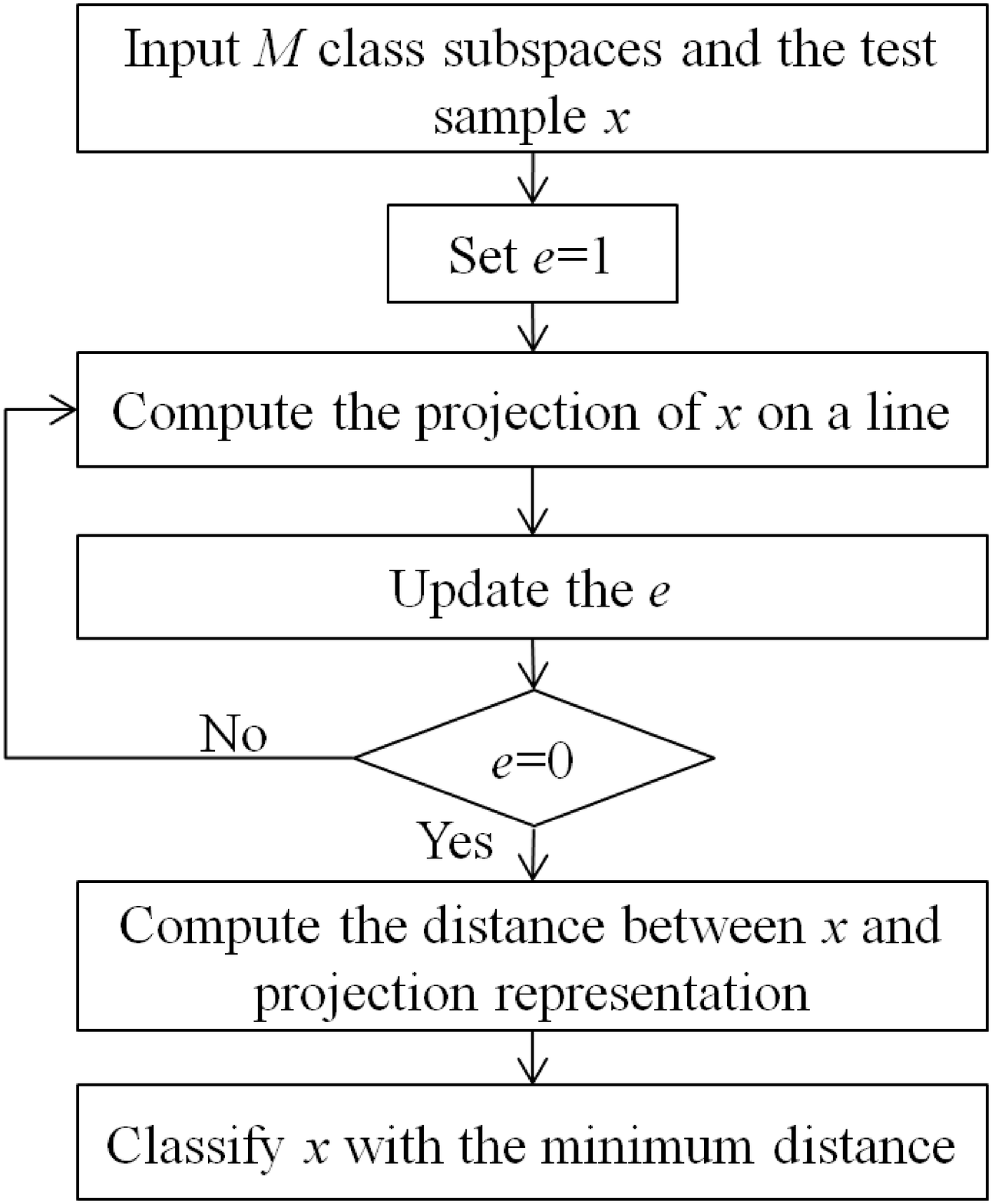}
\end{center}
\vspace{-0.1in}
   \caption{The flowchart of the proposed PRC. $'e=0'$ means that PRC satisfies the stop condition. The detailed information of $e$ can be found in the stop condition.}
\label{fig_flowchart}
\end{figure}

Based on the concept of finding the best representation of each class, this
section proposes a new classifier, called projection representation-based
classification (PRC).
According to the proposed math model, PRC may obtain the approximated projection
point by computing the projection point of the test sample to a line linking with a
pair of training samples iteratively. The flowchart of PRC is shown in Figure \ref{fig_flowchart}.

\subsection{Projection Representation}
To find a point extremely close to the ‘ideal representation’
of a sample point, PRC iteratively computes the projection point of the test
sample on a line. The final result will be treated as the projection representation for classification.

\subsubsection{Start the iteration}

For the first iteration, suppose that a class model $X_c^0$ is described as follows
\begin{eqnarray}
X_c^0=[x_1^{c,0}~~x_2^{c,0}~~\cdots~~x_{N_c}^{c,0}]\in R^{q\times N_c},
\end{eqnarray}
where $x_i^{c,0}=x_i^c$, $i=1,2,\cdots,N_c$. We then select the nearest point $x_{i^*}^{c,0}$ from the class model $X_c^0$ as follow.
\begin{eqnarray}
i^*=\arg\min(\|x-x_i^{c,0}\|),~~i=1,2,\cdots,N_c
\end{eqnarray}
Use the nearest point $x_{i^*}^{c,0}$ and a training sample $x_i^{c,0}$, $i\neq i^*$
to form a line $\overline{x_{i^*}^{c,0}x_i^{c,0}}$. Next, the projection point of the test
sample $x$ on the line $\overline{x_{i^*}^{c,0}x_i^{c,0}}$ can be computed by:

\begin{eqnarray}
p_{i^*}^{c,0}=x_{i^*}^{c,0}+t^0(x_i^{c,0}-x_{i^*}^{c,0})
\end{eqnarray}
where $t\in R$ is the position parameter. The vector $\overline{xp_{i^*}^{c,0}}$
is orthogonal to $\overline{x_{i^*}^{c,0}x_i^{c,0}}$, that is,
$(x-p_{i^*}^{c,0})(x_i^{c,0}-x_{i^*}^{c,0})=0$  where ‘$\bullet$’ denotes the
dot product. Therefore, the position parameter can be computed as
\begin{eqnarray}
t^0=\frac{(x-x_{i^*}^{c,0})^T(x_i^{c,0}-x_{i^*}^{c,0})}{(x_i^{c,0}-x_{i^*}^{c,0})^T(x_i^{c,0}-x_{i^*}^{c,0})}
\end{eqnarray}

\begin{figure}[t]
\begin{center}
\includegraphics[width=1.0\linewidth]{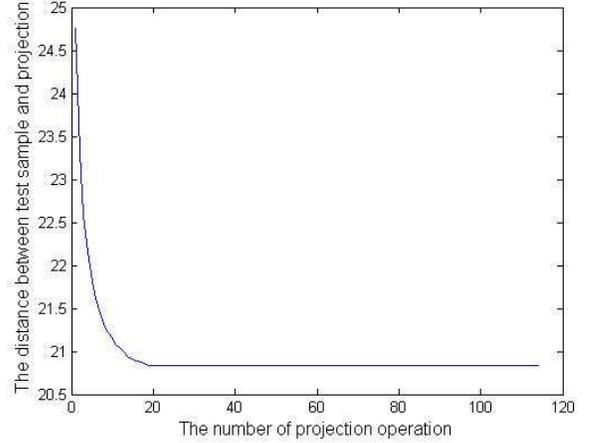}
\end{center}
\vspace{-0.1in}
   \caption{An example of convergence analysis: The distance’s variations between the test sample and approximation projection point in the iteration procedure.}
\label{fig_converge}
\end{figure}

For the $(k+1)$th ($k\geq1$) iteration, it is easy to know that the projection point
$p_{i^*}^{c,k-1}$  in the $k^{th}$ iteration is nearest point. That is, $x_{i^*}^{c,k}=p_{i^*}^{c,k-1}$. The projection point
$p_{i^*}^{c,k}$  in the $(k+1)^{th}$ iteration will be computed by $x$ and a line constituted
by  $p_{i^*}^{c,k-1}$ and another train sample. The procedure of computing the
projection point of $x$ on a line is similar to the first iteration.
\vspace{+0.1in}
\\
\textit{Rule:} All the projection procedure is similar. However, they need to satisfy the
following rule. The number of samples from the class subspace is fixed. The new
projection points will replace the farther point of the line because they are closer to the test sample. All
the samples of class subspace will be sequentially used to constitute the line (as farther point of the line) so that the projection point may contain the information
of all the training samples.

\subsubsection{Convergence Analysis and Stop Condition}

Because the number of iterations is unlimited, we need to determine the condition
for stopping the iteration processes. In order to obtain a good parameter for ending
of the process, an example is given as follows. The training set and test sample are
produced randomly, the dimension of each sample is 5000, and the number of training
samples is 20. Fig. \ref{fig_converge} shows that the distance between the test sample and the
projection point changes with the number of iterations.  As can be seen, the
difference between two adjacent distances tends to zeros. Thus, the stop conditions
of the iteration process are described as follows.\vspace{+0.1in}

 \textit{Condition 1:} Suppose that $p_{i^*}^{c,k-1}$ and $p_{i^*}^{c,k}$ are two nearest
projection points in the $k$th and $(k+1)$th iterations. If $\delta<\delta_0$, the iteration
process stops, where $\delta_0$  is given before the iteration and the threshold
value $\delta$ can be computed as
\begin{eqnarray}
\delta=abs\left(\frac{\|x-p_{i^*}^{c,k-1}\|-\|x-p_{i^*}^{c,k}\|}{\|x-p_{i^*}^{c,k-1}\|+\|x-p_{i^*}^{c,k}\|}   \right).
\end{eqnarray}
Besides, in order to avoid the unpredicted situation, another condition is described as follows.\vspace{+0.1in}
\\
\textit{Condition 2:} Set the maximum iterative times $J$. Based on the Figure \ref{fig_converge}, we
suggest that $J$ is set no more than 100. Notice that this condition is rarely used. It can be treated as an insurance.
\vspace{+0.1in}
\\
Set a stop parameter $e=1$, if one of the two stop conditions is satisfied, $e=0$, the iteration stops. Then the projection representation $p^c$ can be described as
\begin{eqnarray}\label{eq_pr}
p^c=p_{i^*}^{c,k}=x_{i^*}^{c,k}+t^k(x_i^{c,k}-x_{i^*}^{c,k})
\end{eqnarray}
where
\begin{eqnarray}
t^k=\frac{(x-x_{i^*}^{c,k})^T(x_i^{c,k}-x_{i^*}^{c,k})}{(x_i^{c,k}-x_{i^*}^{c,k})^T(x_i^{c,k}-x_{i^*}^{c,k})}
\end{eqnarray}
\\
Notice: For the example of convergence analysis in Figure \ref{fig_converge}, we repeat the experiment
more than one hundred times. The tendency of the distance variations is similar. Select only some valuable samples that is helpful for classification.


\subsubsection{Classification}

Using the Algorithm 1, the approximation projection $p^c$ is obtained for the $c^{th}$ class subspace. The distance between the test sample and the $c^{th}$ class subspace can be computed as
\begin{eqnarray}
d_c(x)=\|x-p^c\|.
\end{eqnarray}
PRC selects the class with the minimum distance
\begin{eqnarray}
\min \limits_{c^*}~d_c(x), c=1,2,\cdots,M.
\end{eqnarray}

\begin{algorithm} [t]
\small
\vspace{0.0in} \caption{Projection Representation}
\begin{description}
  \item[Inputs] The entire training samples $x_i^c$, $c=1,2,\cdots,M$, $i=1,2,\cdots,N_c$ and a test image vector $x\in R^{q\times 1}$. The stop parameter $\delta_0$ and $J$.
  \item[Output] The projection representation $p^c$.
\end{description}
\begin{enumerate}
  \item Set $e=1$; $J=100$; ${\delta _0} = 0.01$

  \textbf{Repeat}

  \item Find the nearest point $x_n^{c,k}$  from the class-models $X_c^k$
  \item Compute the projection point $p_{i*}^{c,k}$ of the test sample $x$ on the line $\overline{x_{i^*}^{c,k}x_i^{c,k}}$ ($i=1,2,\cdots,N_c$ and $i\neq i^*$) as \vspace{-0.05in}
      \begin{eqnarray*}
      \left\{ {\begin{array}{*{20}{c}}
{p_{i*}^{c,k} = x_{i*}^{c,k} + t^k(x_i^{c,k} - x_{{\rm{i*}}}^{c,k})}\\
{t = \frac{{{{(x - x_{i*}^{c,k})}^T}(x_i^{c,k} - x_{i*}^{c,k})}}{{{{(x_i^{c,k} - x_{i*}^{c,k})}^T}(x_i^{c,k} - x_{i*}^{c,k})}}}
\end{array}} \right.
      \end{eqnarray*}\vspace{-0.15in}
  \item  Update the class-models $X_c^k$ using the $p_{i^*}^{c,k}$  to replace the farther point $x_{i}^{c,k}$ of line $\overline{x_{i^*}^{c,k}x_i^{c,k}}$ as $X_c^{k+1}$
  \item Update the parameters $\delta$   by using the projection point $p_{i^*}^{c,k-1}$  in the last iteration and $p_{i^*}^{c,k}$ in this iteration as  \vspace{-0.05in}
  \begin{eqnarray*}
  \delta=abs\left(\frac{\|x-p_{i^*}^{c,k-1}\|-\|x-p_{i^*}^{c,k}\|}{\|x-p_{i^*}^{c,k-1}\|+\|x-p_{i^*}^{c,k}\|}\right)
  \end{eqnarray*}\vspace{-0.15in}
  \item Update the parameter $J=J-1$.
  \item Update the $e$ as\\
  ~~~~~~~~If ($\delta<\delta_0\|J<0$)\\
  ~~~~~~~~~~~~~~~~$e=0$; break;\\
  ~~~~~~~~end if \\
  \textbf{Until the $e=0$ and output the $p^c=p_{i*}^{c,k}$}
\end{enumerate}
\end{algorithm}

%

\subsection{Computational Complex}

Suppose the dimensional of each sample is $q$, it is easy to know that the computational cost of each projection operation is  $O(q)$.
Therefore, the computational complex of PRC is $O(Kq)$ , $K$ is the number of projection operations. From the Figure \ref{fig_converge},
 we know that the iteration number is not large, that is to say, the computational cost of PRC is small.

\section{Proposed DPRC}

PRC obtains the 'ideal projection' while it doesn't use discriminant analysis for classification. Thus, this section pay attention to utilize the labeled training samples set to constitute a more reliable subspace such that the effective discriminant information can be used for classification. In order to obtain an effective discriminant subspace for PRC, we propose a novel method, called discriminant PRC (DPRC), which obtains the discriminant information by maximizing the ratio of the between-class reconstruction error over the within-class reconstruction error by the PRC.

\subsection{Optimization of DPRC}

The proposed DPRC approach is formulated as the optimization problem to maximize the objective function as,
\begin{eqnarray}
\mathop {\max }\limits_P J(P) = \mathop {\max }\limits_P \frac{{{J_b}}}{{{J_w}}}
\end{eqnarray}
where $P$ is the optimal projection matrix that we want to estimate, $J_b$  and $J_w$ denote the between-class and within-class  reconstruction representative metrics, respectively. Then, the goal of the DPRC approach becomes to find an optimal mapping matrix, ${P^{}} = [{p_1},...,{p_k},...,{p_d}]$ which could project the original sample $x_i$  to a new data sample as  ${w_i} = {P^T}{x_i}$ for $i = 1, 2... L$. The proposed projection reduces the dimension and is effective for classification. The above objective function can be also expressed as
\begin{equation}
\begin{split}
J(P) =& \frac{{{J_b}}}{{{J_w}}}\\
 =& \frac{{\frac{1}{{L(M - 1)}}\sum\limits_{i = 1}^L {\sum\limits_{j = 1,j \ne l({x_i})}^M {||{w_i} - w_{ij}^b||} } }}{{\frac{1}{L}\sum\limits_{i = 1}^L {||{w_i} - w_i^w||} }}\\
 =& \frac{{\frac{1}{{L(M - 1)}}\sum\limits_{i = 1}^L {\sum\limits_{j = 1,j \ne l({x_i})}^M {||{P^T}{x_i} - {P^T}x_{ij}^b||} } }}{{\frac{1}{L}\sum\limits_{i = 1}^L {||{P^T}{x_i} - {P^T}x_i^w||} }}
\end{split}
\end{equation}
where $w_{ij}^b = {P^T}x_{ij}^b$ , $w_{ij}^w = {P^T}x_{i}^w$,  $x_{ij}^b$ and $x_{i}^w$  are the between-class and within-class projection vectors. That is, they are projection representation of $x_i$ on $X_j^b$ and $X_i^w$, respectively. They can be calculated by Algorithm 1 with the $x_i$, $X_i^w$ and $X_j^b$.   $X_i^w$ denotes the $l(x_i)$-th class-model in (1)  (don't include the sample $x_i$), $l(x_i)$ denotes the class label of $x_i$ , $X_j^b$ denotes the $j$-th ($j \ne l({x_i})$ ) class-model, With some algebraic derivations in matrices, we have
\begin{equation}
\begin{split}
J(P) =& \frac{{\frac{1}{L}\sum\limits_{i = 1}^L {\sum\limits_{j = 1,j \ne l({x_i})}^M {tr[{P^T}({x_i} - x_{ij}^b){{({x_i} - x_{ij}^b)}^T}P]} } }}{{\frac{1}{L}\sum\limits_{i = 1}^L {tr[{P^T}({x_i} - x_i^w){{({x_i} - x_i^w)}^T}P]} }}\\
 =& tr(\frac{{{P^T}{J_b}P}}{{{P^T}{J_w}P}})
\end{split}
\end{equation}
where
\begin{equation}
{J_b} = \frac{1}{L}\sum\limits_{i = 1}^L {\sum\limits_{j = 1,j \ne l({x_i})}^M {({x_i} - x_{ij}^b){{({x_i} - x_{ij}^b)}^T}} }
\end{equation}
and
\begin{equation}
{J_w} = \frac{1}{L}\sum\limits_{i = 1}^L {({x_i} - x_i^w){{({x_i} - x_i^w)}^T}}
\end{equation}
Afterwards, the objective function can be expressed as
\begin{equation}
\begin{array}{l}
\mathop {\arg \max }\limits_P \frac{{{P^T}{J_b}{P^{}}}}{{{P^T}{J_w}{P^{}}}}\\
\begin{array}{*{20}{c}}
{s.t.}&{{P^T}P = I}
\end{array}
\end{array}
\end{equation}
In order to address the typical small sample size problem, the term $\varepsilon I$ is increased without affecting the subspace. Thus, the objective function can be rewritten as
\begin{equation}
\begin{array}{l}
\mathop {\arg \max }\limits_P \frac{{{P^T}{J_b}{P^{}}}}{{{P^T}({J_w} + \varepsilon I){P^{}}}}\\
\begin{array}{*{20}{c}}
{s.t.}&{{P^T}P = I}
\end{array}
\end{array}
\end{equation}
where $\varepsilon$  is a small number and $I$ is an identity matrix. By using Lagrange multiplier, the projection matrix ${P^{}} = [{p_1},...,{p_k},...,{p_d}]$  that maximizes the objective function, which can be gained by solving the eigen decomposition problem of $\frac{{{J_b}}}{{{J_w} + \varepsilon I}}$  as
\begin{equation}
\begin{array}{*{20}{c}}
{{J_b}{p_k} = {\lambda _k}({J_w} + \varepsilon I){p_k}}&{,k = 1,2,...,d}
\end{array}
\end{equation}
where ${\lambda _1} \ge ... \ge {\lambda _k} \ge ... \ge {\lambda _d}$ is $d$ largest eigenvalues and their corresponding eigenvectors, ${p_1},...,{p_k},...,{p_d}$ of $\frac{{{J_b}}}{{{J_w}}}$. It is noted that ${P^{}} = [{p_1},...,{p_k},...,{p_d}]$  is a $q\times d$ projection matrix, which can project the original $q$-element data vector to the new $d$-element data vector as  ${w_i} = {P^T}{x_i}$ for $i = 1, 2... L$.

\subsection{Classification}

In the above Section, DPRC obtains the effective discriminant space $W=\{w_i\in R^{d\times 1},i=1,2,\cdots,L\}$. Using the discriminant space $W$ and Algorithm 1, the approximation projection $p^c$ is obtained for the $c^{th}$ class subspace. The distance between the test sample and the $c^{th}$ class subspace can be computed as
\begin{eqnarray}
d_c(w)=\|w-p^c\|.
\end{eqnarray}
DPRC selects the class with the minimum distance
\begin{eqnarray}
\min \limits_{c^*}~d_c(w), c=1,2,\cdots,M.
\end{eqnarray}
where ${w} = {P^T}{x}$.

\subsection{DPRC vs ULDA }

This section compares DPRC with a discriminant-based method: Uncertain LDA (ULDA) \cite{UncLDA}. To better explain it, their similarity and difference are given as follows.
\begin{itemize}
  \item Similarity: They both maximize the following objective function $\mathop {\max }\limits_P J(P) = \mathop {\max }\limits_P \frac{{{J_b}}}{{{J_w}}}$, where $J_b, J_w$ are the within-class and between-class scatters. This objective function is the same to that in LDA \cite{LDA}.
  \item Difference: In ULDA, $J_b=S_b+U_b, J_w=S_w+U_w$, where $S_b, S_w$ are the within-class and between-class scatters in LDA. ULDA proposes the uncertain within-class and between-class scatters $U_b, U_w$. In DPRC, the $J_b, J_w$ can be treated as new projection-based within-class and between-class scatters, which has significantly difference to $S_b, S_w$ and $U_b, U_w$.
\end{itemize}

\section{Experimental Results}

This section evaluate the proposed PRC and DPRC on several vision recognition databases.

\subsection{Face recognition}

LFW-a database \cite{LFW} is used in this experiment.
Following \cite{MCT}, we apply 158 subjects that have no less than ten samples for evaluation. The experiment set:
5 samples are randomly selected to form the training set, while other 2 samples are exploited for testing.
The SRC \cite{SRC2}, SVM \cite{SVM}, FDDL \cite{FDDL}, MCT \cite {MCT}, RCR \cite{RCR}, ULDA \cite{UncLDA}, ProCRC \cite{ProCRC} and CRC \cite{CRC} algorithms are chosen for comparison.
Table \ref{t_lfw} illustrates the comparison results of all methods. DPRC obtains better performance than PRC. Compared to the exsiting methods, DPRC has more than 3\% improvement..

\begin{table} [htp]
\caption{The recognition rate (RR) of several classifiers on LFW face database} \label{t_lfw}
\begin{center}
\begin{tabular}{|l|c|l|c|}
\hline
Classifier & Accuracy & Classifier & Accuracy (\%) \\
\hline\hline
SRC  & 44.10 & CRC  & 44.30 \\
SVM  & 43.30 & ULDA  & 44.30 \\
FDDL  & 42.00 & ProCRC  & 44.90 \\
MCT  & 44.90 & \bf PRC  & \bf 46.84 \\
RCR  & 36.70 & \bf DPRC & \bf 47.90 \\
\hline
\end{tabular}
\end{center}
\end{table}


\subsection{Scene classification}

The well-known 15 scene database contains 4,485 images of 15 scene categories \cite{KSPM}. Each image is transformed to spatial pyramid feature provided by \cite{LC-KSVD}.
 The following experimental protocol is used \cite{LLKNNC}: 100 images per class are randomly chosen for training and the rest images are used for testing.
 The D-KSVD \cite {D-KSVD}, LLC \cite{LLC}, LC-KSVD \cite{LC-KSVD}, ULDA \cite{UncLDA},
 LLNMC \cite{LLKNNC}, LLKNNC \cite{LLKNNC}, LRC \cite{LRC}, CRC \cite{CRC}, SRC \cite {SRC2}, ProCRC \cite{ProCRC}, DADL \cite{DADL} methods are chosen for comparison.
 The average classification rate of 10 runs is used to evaluate all methods. From the results in Table \ref{t_15scenes}, our proposed PRC and DPRC obtain the best performance compared with other methods.

\begin{table} [htp]
\begin{center}
\begin{tabular}{|l|c|l|c|}
\hline
Classifier & Accuracy & Classifier & Accuracy (\%) \\
\hline\hline
LRC        & 95.51 & CRC        & 95.95 \\
LLC       &  80.57  & SSRC        & 96.45     \\
D-KSVD      & 89.10 &  SRC       & 96.53\\
LC-KSVD     & 90.40 & ProCRC  & 96.54  \\
ULDA     & 97.70 & DADL  & 98.30   \\
LLKNNC      & 93.54 & \bf PRC          & \bf 98.47  \\
LLNMC     & 97.45 & \bf DPRC          & \bf 98.70 \\
\hline
\end{tabular}
\end{center}
\vspace{-0.1in}
\caption{The recognition rate (RR) of several classifiers on the 15 scenes database.}\label{t_15scenes}
\end{table}

\subsection{Object Classification}

The Caltech101 dataset \cite{Caltech101} has 9,144 images with 102 classes. Following the common experimental settings, we train on 5 samples per class and the rest images are
 used as the testing set. In the experiment, we utilize the 3000-dimension spatial pyramid feature provided by \cite{LC-KSVD} to represent the object image.
  The DNNC \cite{DNNC}, SVM \cite{SVM}, FDDL  \cite{FDDL}, D-KSVD \cite {D-KSVD},
  LRC \cite{LRC}, CRC \cite{CRC}, SRC \cite {SRC2}, SSRC  \cite{SSRC}, ProCRC \cite{ProCRC} and ULDA \cite{UncLDA} methods are chosen for comparison.
 The experiment results are shown in Table \ref{t_calt101}.
  As can be observed, DPRC gains the best performance compared several popular methods.

\begin{table} [htp]
\begin{center}
\begin{tabular}{|l|c|l|c|}
\hline
Classifier & Accuracy & Classifier & Accuracy(\%) \\
\hline\hline
DNNC     & 46.60 & SVM      & 47.88 \\
SRC      & 48.80 & SSRC    & 47.10 \\
FDDL     & 49.80 & ULDA  & 48.54 \\
D-KSVD    & 49.60 & ProCRC  & 47.80 \\
CRC      & 44.68 & \bf PRC      & \bf 50.66 \\
LRC      & 47.54 & \bf DPRC      & \bf 50.80 \\
\hline
\end{tabular}
\end{center}
\vspace{-0.1in}
\caption{The recognition rate (RR) of several classifiers on the Caltech 101 database.}\label{t_calt101}
\end{table}

\subsection{Action Recognition}
The Ucf50 action dataset \cite{UCF50} has 6,680 action videos with 50 action categories, which was taken from YouTube. For fair comparison,  we follow the ref. \cite{DADL}:
Divide the database into five folds, use four folds for training and one fold for testing.
We use PCA \cite{RPCA-AOM} to reduce the action bank features \cite{actionbank} to 5000 dimensions.
  The CRC \cite{CRC}, SRC \cite {SRC2}, DLSI  \cite{DLSI}, ULDA \cite{UncLDA}, SSRC \cite{SSRC} , FDDL  \cite{FDDL}, LC-KSVD  \cite{LC-KSVD}, DPL  \cite{DPL}, ProCRC \cite{ProCRC} and DADL \cite{DADL} methods are chosen for comparison.
 The experiment results are shown in Table \ref{t_ucf50}.
  DPRC has the better performance than PRC and gains the best performance compared with several popular methods.

  \begin{table} [htp]
\begin{center}
\begin{tabular}{|l|c|l|c|}
\hline
Classifier & Accuracy & Classifier & Accuracy(\%) \\
\hline\hline
CRC     & 75.60 & DPL      & 77.40 \\
SSRC    &  76.40 &     ULDA       &  77.60       \\
SRC      & 75.00 & ProCRC    & 77.40 \\
DLSI     & 75.40 & DADL  & 78.00 \\
FDDL    & 76.50 & \bf PRC  & \bf 78.50 \\
LC-KSVD      & 70.10 & \bf DPRC      & \bf 79.10 \\
\hline
\end{tabular}
\end{center}
\vspace{-0.1in}
\caption{The recognition rate (RR) of several classifiers on the Ucf50 action database.}\label{t_ucf50}
\end{table}

\subsection{Compare with Deep Learning based Methods}
The Caltech-256 dataset \cite{Caltech256} has 30,608 object images of 256 object class, each class has at least 80 object images. To access the performance of PRC and DPRC for object recognition with the deep-learning-based feature, we follow Ref. \cite{NAC}, randomly select 60 images for training, the rest images are used for testing. Five deep learning based methods are used for comparison. They include NAC \cite{NAC}, CNN-S \cite{CNN-M}, ZF  \cite{ZF}, CNN-M  \cite{CNN-M} and VGG19  \cite{VGG19}.
 The experiment results are shown in Table \ref{t_caltech256}. As we can see, the proposed methods with deep feature obtain the better performance than the deep learning based methods. The proposed DPRC has the better performance compared to proposed PRC.

  \begin{table} [h]
\begin{center}
\begin{tabular}{|l|c|}
\hline
Classification Methods & Accuracy (\%)\\
\hline\hline
CNN-S     & 77.6  \\
ZF    & 74.2  \\
CNN-M     & 75.5  \\
VGG19+ SVM    & 83.9  \\
NAC     & 84.1  \\
\hline
\bf VGG19+PRC    & \bf 84.9  \\
\bf VGG19+DPRC     & \bf 85.3  \\
\hline
\end{tabular}
\end{center}
\vspace{-0.1in}
\caption{Accuracy of several methods on the Caltech 256 object database.}\label{t_caltech256}
\end{table}

\section{Conclusion}

In this paper, projection representation-based classification (PRC) has been proposed for image recognition.
The PRC uses the iteratively projection procedures to obtain a point to closely approximate the 'ideal representation'.
The objectives of PRC, SRC and LRC are similar but PRC gains the better representation. Based on PRC, the discriminant PRC (DPRC) is further proposed.
DPRC increase the discriminant information for PRC such that it obtains the better performance.
The experimental results on several well-known databases have confirmed the good performance of the proposed PRC and DPRC for face,
objection, scene and action recognitions. Moreover, PRC and DPRC with deep-learning-based feature can obtain the better performance than deep learning based methods
\\

\section{Acknowledgments}
Thanks for the valuable suggestions of Editor and reviewers.
This work was supported in part by the Macau Science and Technology Development Fund under Grant FDCT/016/2015/A1 and by the Research Committee at University of Macau under Grants MYRG2014-00003-FST and MYRG2016-00123-FST.

%

\bibliography{egbib}

\begin{thebibliography}{}

\bibitem[\protect\citeauthoryear{Basri and Jacobs}{2003}]{LRLS}
Basri, R., and Jacobs, D.~W.
\newblock 2003.
\newblock Lambertian reflectance and linear subspaces.
\newblock {\em TPAMI} 25(2):218--233.

\bibitem[\protect\citeauthoryear{Cai \bgroup et al\mbox.\egroup
  }{2016}]{ProCRC}
Cai, S.; Zhang, L.; Zuo, W.; and Feng, X.
\newblock 2016.
\newblock A probabilistic collaborative representation based approach for
  pattern classification.
\newblock In {\em CVPR},  2950--2959.

\bibitem[\protect\citeauthoryear{Chatfield \bgroup et al\mbox.\egroup
  }{2014}]{CNN-M}
Chatfield, K.; Simonyan, K.; Vedaldi, A.; and Zisserman, A.
\newblock 2014.
\newblock Return of the devil in the details: Delving deep into convolutional
  nets.
\newblock In {\em BMVC},  1--12.

\bibitem[\protect\citeauthoryear{Deng, Hu, and Guo}{2013}]{SSRC}
Deng, W.; Hu, J.; and Guo, J.
\newblock 2013.
\newblock In defense of sparsity based face recognition.
\newblock In {\em CVPR},  399--406.
\newblock IEEE.

\bibitem[\protect\citeauthoryear{Fei-Fei, Fergus, and
  Perona}{2007}]{Caltech101}
Fei-Fei, L.; Fergus, R.; and Perona, P.
\newblock 2007.
\newblock Learning generative visual models from few training examples: An
  incremental bayesian approach tested on 101 object categories.
\newblock {\em Computer Vision and Image Understanding} 106(1):59--70.

\bibitem[\protect\citeauthoryear{Feng and Zhou}{2016a}]{ILRC}
Feng, Q., and Zhou, Y.
\newblock 2016a.
\newblock Iterative linear regression classification for image recognition.
\newblock In {\em ICASSP},  1566--1570.

\bibitem[\protect\citeauthoryear{Feng and Zhou}{2016b}]{KCSR}
Feng, Q., and Zhou, Y.
\newblock 2016b.
\newblock Kernel combined sparse representation for disease recognition.
\newblock {\em TMM} 18(10):1956--1968.

\bibitem[\protect\citeauthoryear{Feng and Zhou}{2017}]{KRDU}
Feng, Q., and Zhou, Y.
\newblock 2017.
\newblock Kernel regularized data uncertainty for action recognition.
\newblock {\em TCSVT} 27(3):577--588.

\bibitem[\protect\citeauthoryear{Feng, Zhou, and Lan}{2016}]{PLRC}
Feng, Q.; Zhou, Y.; and Lan, R.
\newblock 2016.
\newblock Pairwise linear regression classification for image set retrieval.
\newblock In {\em CVPR},  4865--4872.

\bibitem[\protect\citeauthoryear{Griffin, Holub, and Perona}{2007}]{Caltech256}
Griffin, G.; Holub, A.; and Perona, P.
\newblock 2007.
\newblock Caltech-256 object category dataset.

\bibitem[\protect\citeauthoryear{Gu \bgroup et al\mbox.\egroup }{2014}]{DPL}
Gu, S.; Zhang, L.; Zuo, W.; and Feng, X.
\newblock 2014.
\newblock Projective dictionary pair learning for pattern classification.
\newblock In {\em NIPS},  793--801.

\bibitem[\protect\citeauthoryear{Guo \bgroup et al\mbox.\egroup }{2016}]{DADL}
Guo, J.; Guo, Y.; Kong, X.; Zhang, M.; and He, R.
\newblock 2016.
\newblock Discriminative analysis dictionary learning.
\newblock In {\em AAAI},  1617--1623.

\bibitem[\protect\citeauthoryear{Haeb-Umbach and Ney}{1992}]{LDA}
Haeb-Umbach, R., and Ney, H.
\newblock 1992.
\newblock Linear discriminant analysis for improved large vocabulary continuous
  speech recognition.
\newblock In {\em ICASSP}, volume~1,  13--16.

\bibitem[\protect\citeauthoryear{Jiang, Lin, and Davis}{2013}]{LC-KSVD}
Jiang, Z.; Lin, Z.; and Davis, L.~S.
\newblock 2013.
\newblock Label consistent k-svd: Learning a discriminative dictionary for
  recognition.
\newblock {\em TPAMI} 35(11):2651--2664.

\bibitem[\protect\citeauthoryear{Lazebnik, Schmid, and Ponce}{2006}]{KSPM}
Lazebnik, S.; Schmid, C.; and Ponce, J.
\newblock 2006.
\newblock Beyond bags of features: Spatial pyramid matching for recognizing
  natural scene categories.
\newblock In {\em CVPR}, volume~2,  2169--2178.
\newblock IEEE.

\bibitem[\protect\citeauthoryear{Liu and Liu}{2015}]{LLKNNC}
Liu, Q., and Liu, C.
\newblock 2015.
\newblock A novel locally linear knn model for visual recognition.
\newblock In {\em CVPR},  1329--1337.

\bibitem[\protect\citeauthoryear{Luo \bgroup et al\mbox.\egroup
  }{2016}]{RPCA-AOM}
Luo, M.; Nie, F.; Chang, X.; Yang, Y.; Hauptmann, A.; and Zheng, Q.
\newblock 2016.
\newblock Avoiding optimal mean robust pca/2dpca with non-greedy l1-norm
  maximization.
\newblock In {\em IJCAI},  1802--1808.

\bibitem[\protect\citeauthoryear{Naseem, Togneri, and Bennamoun}{2010}]{LRC}
Naseem, I.; Togneri, R.; and Bennamoun, M.
\newblock 2010.
\newblock Linear regression for face recognition.
\newblock {\em TPAMI} 32(11):2106--2112.

\bibitem[\protect\citeauthoryear{Ramirez, Sprechmann, and Sapiro}{2010}]{DLSI}
Ramirez, I.; Sprechmann, P.; and Sapiro, G.
\newblock 2010.
\newblock Classification and clustering via dictionary learning with structured
  incoherence and shared features.
\newblock In {\em CVPR},  3501--3508.
\newblock IEEE.

\bibitem[\protect\citeauthoryear{Reddy and Shah}{2013}]{UCF50}
Reddy, K.~K., and Shah, M.
\newblock 2013.
\newblock Recognizing 50 human action categories of web videos.
\newblock {\em Machine Vision and Applications} 24(5):971--981.

\bibitem[\protect\citeauthoryear{Sadanand and Corso}{2012}]{actionbank}
Sadanand, S., and Corso, J.~J.
\newblock 2012.
\newblock Action bank: A high-level representation of activity in video.
\newblock In {\em CVPR},  1234--1241.

\bibitem[\protect\citeauthoryear{Saeidi, Astudillo, and Kolossa}{2016}]{UncLDA}
Saeidi, R.; Astudillo, R.~F.; and Kolossa, D.
\newblock 2016.
\newblock Uncertain lda: Including observation uncertainties in discriminative
  transforms.
\newblock {\em TPAMI} 38(7):1479--1488.

\bibitem[\protect\citeauthoryear{Sch{\"u}ldt, Laptev, and Caputo}{2004}]{SVM}
Sch{\"u}ldt, C.; Laptev, I.; and Caputo, B.
\newblock 2004.
\newblock Recognizing human actions: a local svm approach.
\newblock In {\em ICPR}, volume~3,  32--36.

\bibitem[\protect\citeauthoryear{Simon and Rodner}{2015}]{NAC}
Simon, M., and Rodner, E.
\newblock 2015.
\newblock Neural activation constellations: Unsupervised part model discovery
  with convolutional networks.
\newblock In {\em ICCV},  1143--1151.

\bibitem[\protect\citeauthoryear{Simonyan and Zisserman}{2014}]{VGG19}
Simonyan, K., and Zisserman, A.
\newblock 2014.
\newblock Very deep convolutional networks for large-scale image recognition.
\newblock {\em arXiv:1409.1556}.

\bibitem[\protect\citeauthoryear{Wang \bgroup et al\mbox.\egroup }{2010}]{LLC}
Wang, J.; Yang, J.; Yu, K.; Lv, F.; Huang, T.; and Gong, Y.
\newblock 2010.
\newblock Locality-constrained linear coding for image classification.
\newblock In {\em CVPR},  3360--3367.

\bibitem[\protect\citeauthoryear{Wright \bgroup et al\mbox.\egroup
  }{2009}]{SRC2}
Wright, J.; Yang, A.~Y.; Ganesh, A.; Sastry, S.~S.; and Ma, Y.
\newblock 2009.
\newblock Robust face recognition via sparse representation.
\newblock {\em TPAMI} 31(2):210--227.

\bibitem[\protect\citeauthoryear{Yang \bgroup et al\mbox.\egroup }{2012}]{RCR}
Yang, M.; Zhang, L.; Zhang, D.; and Wang, S.
\newblock 2012.
\newblock Relaxed collaborative representation for pattern classification.
\newblock In {\em CVPR},  2224--2231.

\bibitem[\protect\citeauthoryear{Yang \bgroup et al\mbox.\egroup }{2014}]{FDDL}
Yang, M.; Zhang, L.; Feng, X.; and Zhang, D.
\newblock 2014.
\newblock Sparse representation based fisher discrimination dictionary learning
  for image classification.
\newblock {\em IJCV} 109(3):209--232.

\bibitem[\protect\citeauthoryear{Zeiler and Fergus}{2014}]{ZF}
Zeiler, M.~D., and Fergus, R.
\newblock 2014.
\newblock Visualizing and understanding convolutional networks.
\newblock In {\em ECCV},  818--833.

\bibitem[\protect\citeauthoryear{Zhang and Li}{2010}]{D-KSVD}
Zhang, Q., and Li, B.
\newblock 2010.
\newblock Discriminative k-svd for dictionary learning in face recognition.
\newblock In {\em CVPR},  2691--2698.

\bibitem[\protect\citeauthoryear{Zhang \bgroup et al\mbox.\egroup
  }{2006}]{DNNC}
Zhang, H.; Berg, A.~C.; Maire, M.; and Malik, J.
\newblock 2006.
\newblock Svm-knn: Discriminative nearest neighbor classification for visual
  category recognition.
\newblock In {\em CVPR}, volume~2,  2126--2136.

\bibitem[\protect\citeauthoryear{Zhang \bgroup et al\mbox.\egroup }{2015}]{MCT}
Zhang, B.; Perina, A.; Murino, V.; and Del~Bue, A.
\newblock 2015.
\newblock Sparse representation classification with manifold constraints
  transfer.
\newblock In {\em CVPR},  4557--4565.

\bibitem[\protect\citeauthoryear{Zhang, Yang, and Feng}{2011}]{CRC}
Zhang, L.; Yang, M.; and Feng, X.
\newblock 2011.
\newblock Sparse representation or collaborative representation: Which helps
  face recognition?
\newblock In {\em ICCV},  471--478.

\bibitem[\protect\citeauthoryear{Zhu \bgroup et al\mbox.\egroup }{2012}]{LFW}
Zhu, P.; Zhang, L.; Hu, Q.; and Shiu, S.~C.
\newblock 2012.
\newblock Multi-scale patch based collaborative representation for face
  recognition with margin distribution optimization.
\newblock In {\em ECCV}.
\newblock  822--835.

\end{thebibliography}
\bibliographystyle{aaai}

\end{document}